\documentclass{article}
\pdfoutput=1
\usepackage{blindtext}
\usepackage{graphicx}
\usepackage{amsmath}
\usepackage{amssymb}
\usepackage{array}
\usepackage{subfig}
\usepackage{hyperref}
\usepackage{shortcuts}
\usepackage{bm}
\usepackage[ruled,norelsize]{algorithm2e}

\DeclareMathOperator{\Tr}{Tr}

\makeatletter
\newcommand*{\inlineequation}[2][]{%
  \begingroup
    \refstepcounter{equation}%
    \ifx\\#1\\%
    \else
      \label{#1}%
    \fi
    \relpenalty=10000 %
    \binoppenalty=10000 %
    \ensuremath{%
      #2%
    }%
    ~\@eqnnum
  \endgroup
}
\makeatother

\begin{document}

\title{Faster independent component analysis by preconditioning with Hessian approximations}

\author{Pierre Ablin \thanks{P. Ablin works at Inria, Parietal Team, Universit\'e Paris-Saclay, Saclay, France; e-mail: pierre.ablin@inria.fr},
        Jean-Francois Cardoso \thanks{J. F. Cardoso is with the Institut d'Astrophysique de Paris, CNRS, Paris, France; e-mail: cardoso@iap.fr},
        and Alexandre Gramfort 
\thanks{A. Gramfort is with Inria, Parietal Team, Universit\'e Paris-Saclay, Saclay, France; e-mail: alexandre.gramfort@inria.fr}}
   


\maketitle

\begin{abstract}

Independent Component Analysis (ICA) is a technique for unsupervised exploration of multi-channel data that is widely used in observational sciences. In its classic form, ICA relies on modeling the data as linear mixtures of non-Gaussian independent sources.  The maximization of the corresponding likelihood is a challenging problem if it has to be completed quickly and accurately on large sets of real data. We introduce the Preconditioned ICA for Real Data (Picard) algorithm, which is a relative L-BFGS algorithm preconditioned with sparse Hessian approximations. Extensive numerical comparisons to several algorithms of the same class demonstrate the superior performance of the proposed technique, especially on real data, for which the ICA model does not necessarily hold.

\end{abstract}


\section{Introduction}
\label{sec:intro}

Independent Component Analysis (ICA)~\cite{comon1992independent,hyvarinen2000independent} is a multivariate data exploration tool massively used across scientific disciplines such as neuroscience~\cite{Makeig30091997,calhoun-etal:01,beckmann-etal:04,delorme2012independent}, astronomy~\cite{nuzillard2000blind,maino2002all,cadavid2008principal}, chemistry~\cite{vrabie-etal:07,rutledge2013independent} or biology~\cite{lee2003application, scholz2004metabolite}. The underlying assumption of ICA is that the data are obtained by combining latent components which are statistically independent. The linear ICA problem addresses the case where latent variables and observations are linked by a linear transform. Then, ICA boils down to estimating a linear transform of the input signals into `source signals' which are as independent as possible. 


The strength and wide applicability of ICA come from its limited number of assumptions. For ICA to become a well-posed problem it is only required that all sources except one are non-Gaussian and statistically independent~\cite{comon1992independent}. The generality of this concept explains the usefulness of ICA in many domains. 


An early and popular ICA algorithm is Infomax~\cite{bell1995information}.
It is widely used in neuroscience and is distributed in most neural processing toolboxes (e.g. EEGLAB~\cite{delorme2004eeglab} or MNE~\cite{mne}). 
It can be shown to be a maximum likelihood estimator~\cite{cardoso1997infomax} based on a non Gaussian component model.  
However, Infomax  maximizes the likelihood using a stochastic gradient algorithm which may require some hand-tuning and often fails to converge~\cite{montoya2017caveats}, or only converges slowly. 

Since speed is important in data exploration, various methods have been proposed for a faster maximization of the Infomax likelihood by using curvature information, that is by exploiting not only the gradient of the likelihood as in Infomax but also its second derivatives.
We briefly review some of the methods found in the literature.


The most natural way of using curvature is to use the complete set of second derivatives (the Hessian) to set up the Newton method but it faces several difficulties: 
the  Hessian is a large object, costly to evaluate and to invert for large data sets. It also has to be regularized since the ICA likelihood is not convex.  The cost issue is addressed in~\cite{tillet2017infomax} by using a truncated Newton algorithm: an approximate  Newton direction is found by an early stopping (truncation) of its computation via a conjugate gradient method.  Further, each step in this computation is quickly computed by a `Hessian-free' formula.
Another approach to exploit curvature is to use approximations of the Hessian, obtained by assuming that the current signals are independent (see \textit{e.g.}~\cite{pham1997blind, amari1997stability} or section~\ref{sec:methods}). 
For instance, a simple quasi-Newton method is proposed in~\cite{zibulevsky2003blind} and in AMICA~\cite{palmer2008newton}, and a trust-region algorithm in~\cite{choi2007relative}.

We have re-implemented and compared these methods (see section~\ref{sec:expe})
and found that the Hessian approximations do yield a low cost per iteration but that they are not accurate enough on real data (which cannot be expected to follow the ICA model at high accuracy, \textit{e.g.} in presence of some correlated sources). 
The approach investigated in this article overcomes this problem by using an optimization algorithm which `learns' curvature from the past iterations of the solver (L-BFGS~\cite{byrd1995limited}), and accelerates it by preconditioning with Hessian approximations.



This article is organized as follows. 
In section~\ref{sec:methods}, we recall the expression of the gradient and Hessian of the log-likelihood. We show how simple Hessian approximations can be obtained and regularized. That allows the L-BFGS method to be preconditioned at low cost yielding the Preconditioned ICA for Real Data (Picard) algorithm described in section~\ref{sec:prec}.  In section~\ref{sec:otherwork}, we detail related algorithms mentioned in the introduction.
Finally, section~\ref{sec:expe} illustrates the superior behavior of the Picard algorithm by extensive numerical experiments on synthetic signals, on multiple electroencephalography (EEG) datasets, on functional Magnetic Resonance Imaging (fMRI) data and on natural images.

\section{Likelihood and derivatives}
\label{sec:methods}

\subsection*{Notation}
\label{sec:notation}

The Frobenius matrix scalar product is denoted by $\langle M \lvert M' \rangle = \Tr(M^\top M') = \sum_{i,j} M_{ij}M'_{ij}$, and $\| M \| = \sqrt{\langle M \lvert M \rangle}$ is the associated Frobenius matrix norm.
Let $B$ be a fourth order tensor of size $N \times N \times N \times N$.  Its application to a $N \times N$ matrix $M$ is denoted 
$BM$, a $N\times N$ matrix with entries $(BM)_{ij} =\sum_{k, l} B_{ijkl}M_{kl}$. 
We also denote $\langle M' \lvert B \rvert M \rangle = \langle M' \lvert BM \rangle = \sum_{i, j, k, l} B_{ijkl}M'_{ij}M_{kl}$. 
The Kronecker symbol $\delta_{ij}$ equals $1$ for $i = j$ and $0$ otherwise.

The complexity of an operation is said to go as $\Theta(f(N, T))$ for a real function $f$ if there exist two constants $0 < c_1 < c_2$ such that the cost of that operation is in the interval $[ c_1 f(N, T), c_2 f(N, T)]$ for all $T, N$. 

\subsection{Non Gaussian likelihood for ICA}
\label{sec:maxlik}

The ICA likelihood for a data set  $X = [x_1,.., x_N]^\top \in \bbR^{N \times T}$ of $N$ signals $x_1,\dots,x_N$ of length $T$ is based on the linear model $X=AS$
where the $N\times N$ mixing matrix $A$ is unknown and the source matrix $S$ has $N$ statistically independent zero-mean rows.
If each row of $S$ is modeled as i.i.d. with $p_i(\cdot)$ denoting the common distribution of the samples of the $i$th source,  the likelihood of $A$ is then~\cite{pham1997blind}:
\begin{equation}
    p(X \lvert A) = \prod_{t = 1}^{T} \frac{1}{\lvert \det(A) \rvert} \prod_{i = 1}^{N}p_i([A^{-1} x]_i(t)) \enspace .
\end{equation}
It is convenient to work with the negative averaged log-likelihood parametrized by the unmixing matrix $W = A^{-1}$, that is, $\mathcal{L}(W) = -\frac1T \log p(X \lvert W^{-1})$. It is given by:
\begin{equation}
    \mathcal{L}(W) = -\log\lvert\det(W)\rvert -
    \hat{E}\left[\sum_{i = 1}^{N} \log(p_i(y_i(t))\right] \enspace ,
\label{eq:loglik}
\end{equation}
where $\hat{E}$ denotes the empirical mean (sample average) and where, implicitly, $Y=WX$.
Our aim is to minimize $ \mathcal{L}(W)$ with respect to $W$ which amounts to solving the ICA problem in the maximum likelihood sense.

We note from the start that this optimization problem is not convex for a simple reason: if $W^*$ minimizes the objective function, any permutation of the columns of $W^*$ gives another equivalent minimizer. 

In this paper, we focus on fast and accurate minimization of $\mathcal L (W)$ for a given source model, that is, working with fixed predetermined densities $p_i$. It corresponds to the standard Infomax model commonly used in practice.
In particular, our experiments use $-\log(p_i(\cdot)) = 2\log(\cosh(\cdot/ 2)) + \mathrm{cst}$, which is the density model in standard Infomax.

In the following, the \emph{ICA mixture model} is said to hold if the signals actually are a mixture of independent components.  We stress that \emph{on real data}, the ICA mixture model is not expected to hold exactly.

\subsection{Relative variations of the objective function}
\label{sec:objective}

The variation of $\mathcal{L}(W)$ with respect to a \emph{relative variation} of $W$ is described, up to second order, by the Taylor expansion of $\mathcal{L}((I+\mathcal{E})W)$ in terms of a `small' $N\times N$ matrix $\mathcal{E}$:
\begin{equation}
\mathcal{L}((I + \mathcal{E})W) = \mathcal{L}(W) + \langle G \lvert \mathcal{E}\rangle + \frac{1}{2} \langle \mathcal{E} \lvert H \rvert  \mathcal{E} \rangle +  \mathcal{O}(\lvert \lvert \mathcal{E} \rvert \rvert ^3).
\label{eq:taylor}
\end{equation}
The first order term is controlled by the $N\times N$ matrix $G$, called the relative gradient~\cite{cardoso1996equivariant} and the second-order term depends on the $N\times N\times N \times N$ tensor $H$, called the relative Hessian~\cite{zibulevsky2003blind}. 
Both these quantities can be obtained from the second order expansions of $\log\det(\cdot)$ and $\log p_i(\cdot)$:
\begin{align*}
 \log\lvert\det(I + \mathcal{E})\rvert 
 &=
    \Tr(\mathcal{E})
    - \frac{1}{2}\Tr(\mathcal{E}^2)
    + \mathcal{O}(\lvert \lvert \mathcal{E} \rvert \rvert ^3),
  \\
\log(p_i(y+e)) 
&= \log(p_i(y)) - \psi_i(y) e - \frac{1}{2} \psi'_i(y) e^2 + \mathcal{O}(e^3),
\end{align*}
where $\psi_i = -\frac{p_i'}{p_i}$ is called the \emph{score function} (equal to $\tanh(\cdot/2)$ for the standard Infomax density).
Collecting and re-arranging terms yields at first-order the classic expression
\begin{equation}
G_{i j} = \hat{E}[\psi_i(y_i)y_j] -\delta_{i j} \quad\text{or}\quad
G(Y) = \frac{1}{T} \psi(Y) Y^\top - Id
\label{eq:relatgrad}
\end{equation}
and, at second order, the relative Hessian:
\begin{equation}
H_{i j k l} = \delta_{i l} \delta_{j k} + \delta_{ik} \, \hat{E}[\psi_i'(y_i)y_jy_l]  \enspace .
\label{eq:hessian}
\end{equation}


%

%

Note that the relative Hessian is sparse. Indeed, it has only of the order of $N^3$ non-zero coefficients: $ \delta_{i l} \delta_{j k} \neq 0$ for $i=l$ and $j=k$ which corresponds to $N^2$ coefficients, and $\delta_{ik} \neq 0$ for $i = k$ which happens $N^3$ times. 
This means that for a practical application with 100 sources the Hessian easily fits in memory. However, its computation requires the evaluation of the terms $\hat{E}[\psi_i'(y_i)y_jy_l]$, resulting in a $\Theta(N^3 \times T)$ complexity.
This fact and the necessity of regularizing the Hessian (which is not necessarily positive definite) in Newton methods motivate the consideration of Hessian approximations which are faster to compute and easier to regularize.

\subsection{Hessian Approximations}

\begin{figure}[htp]
  \centering
  \includegraphics[width=0.32\columnwidth]{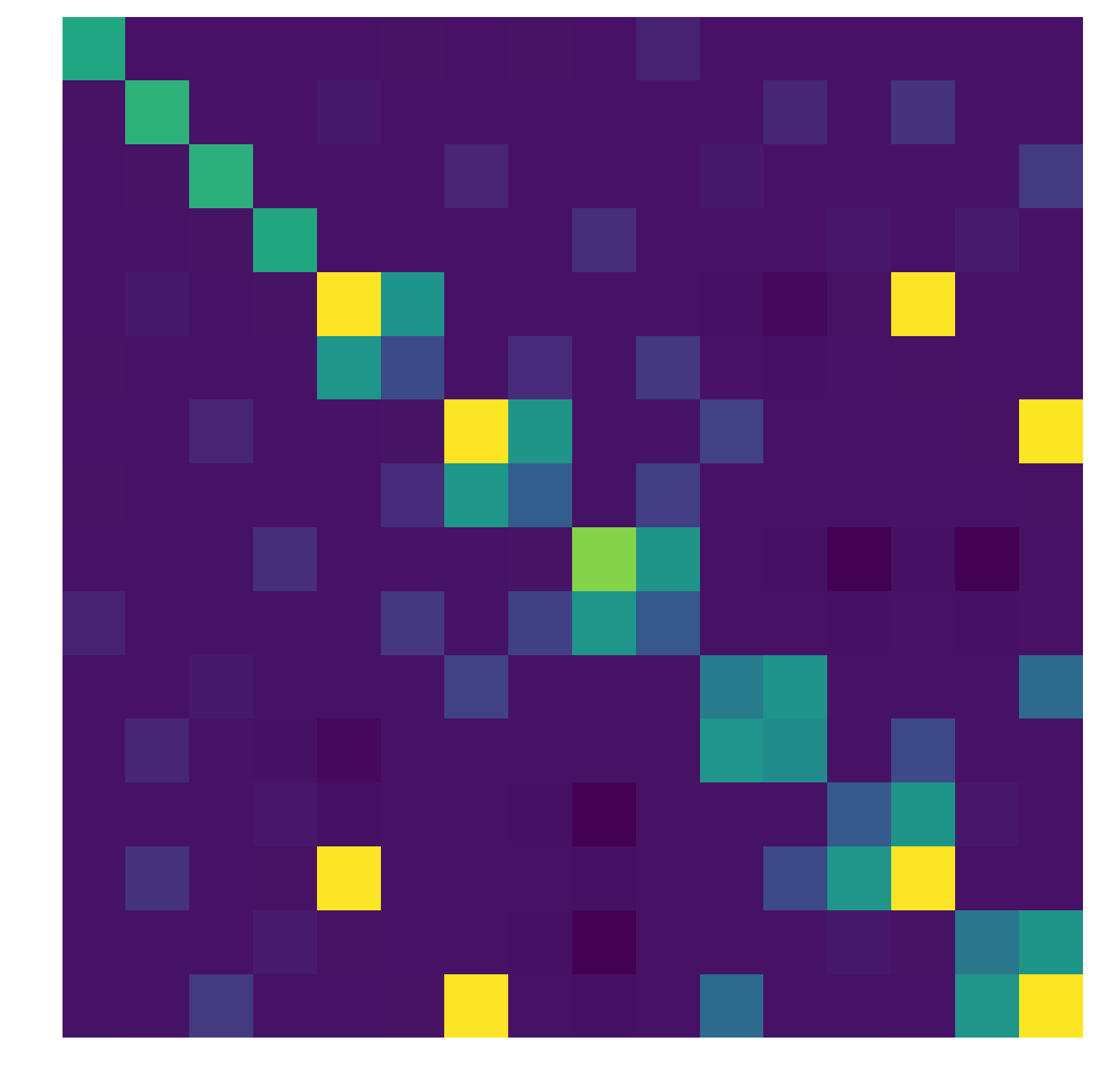}\hfill
  \includegraphics[width=0.32\columnwidth]{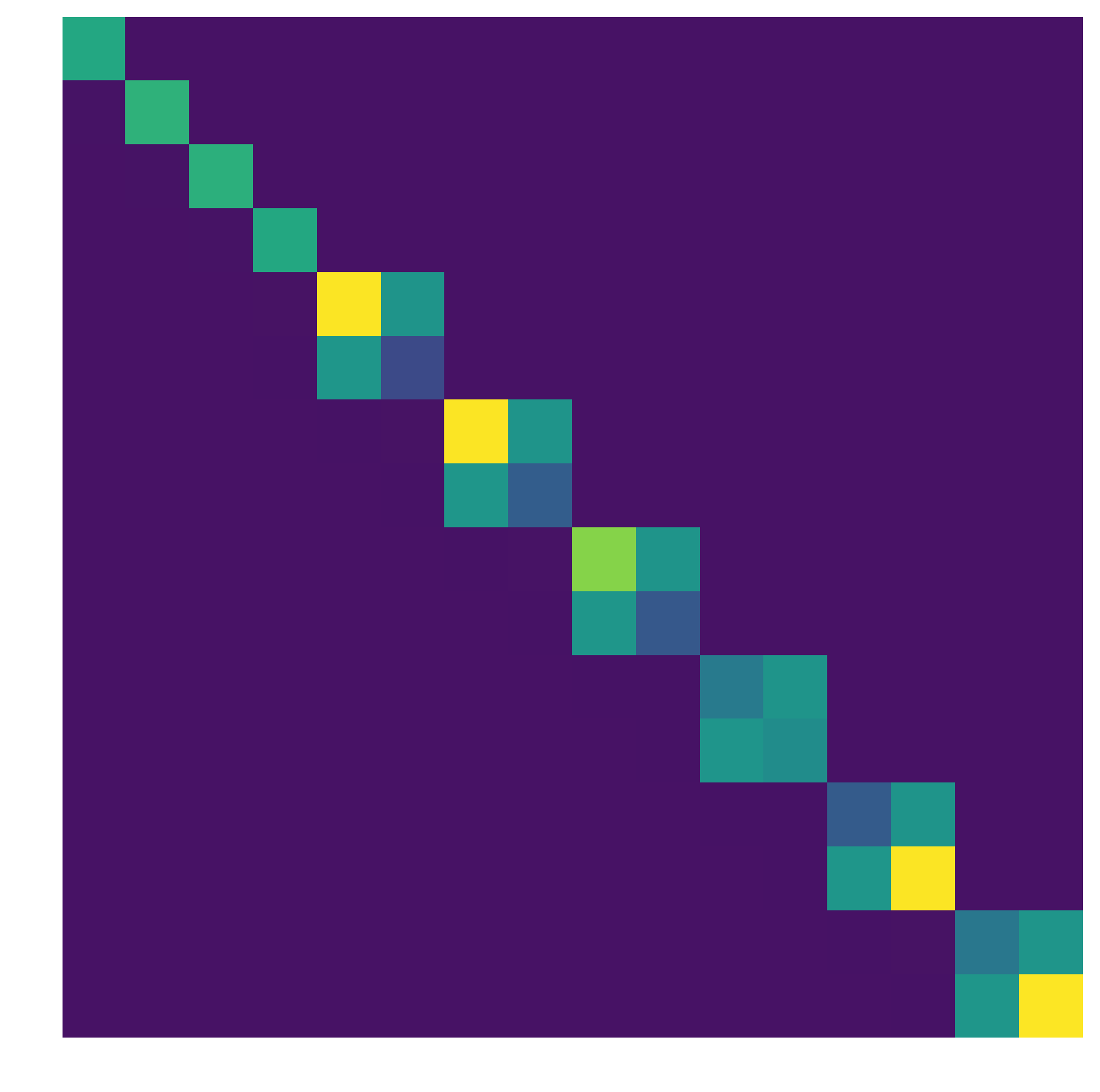}\hfill
  \includegraphics[width=0.32\columnwidth]{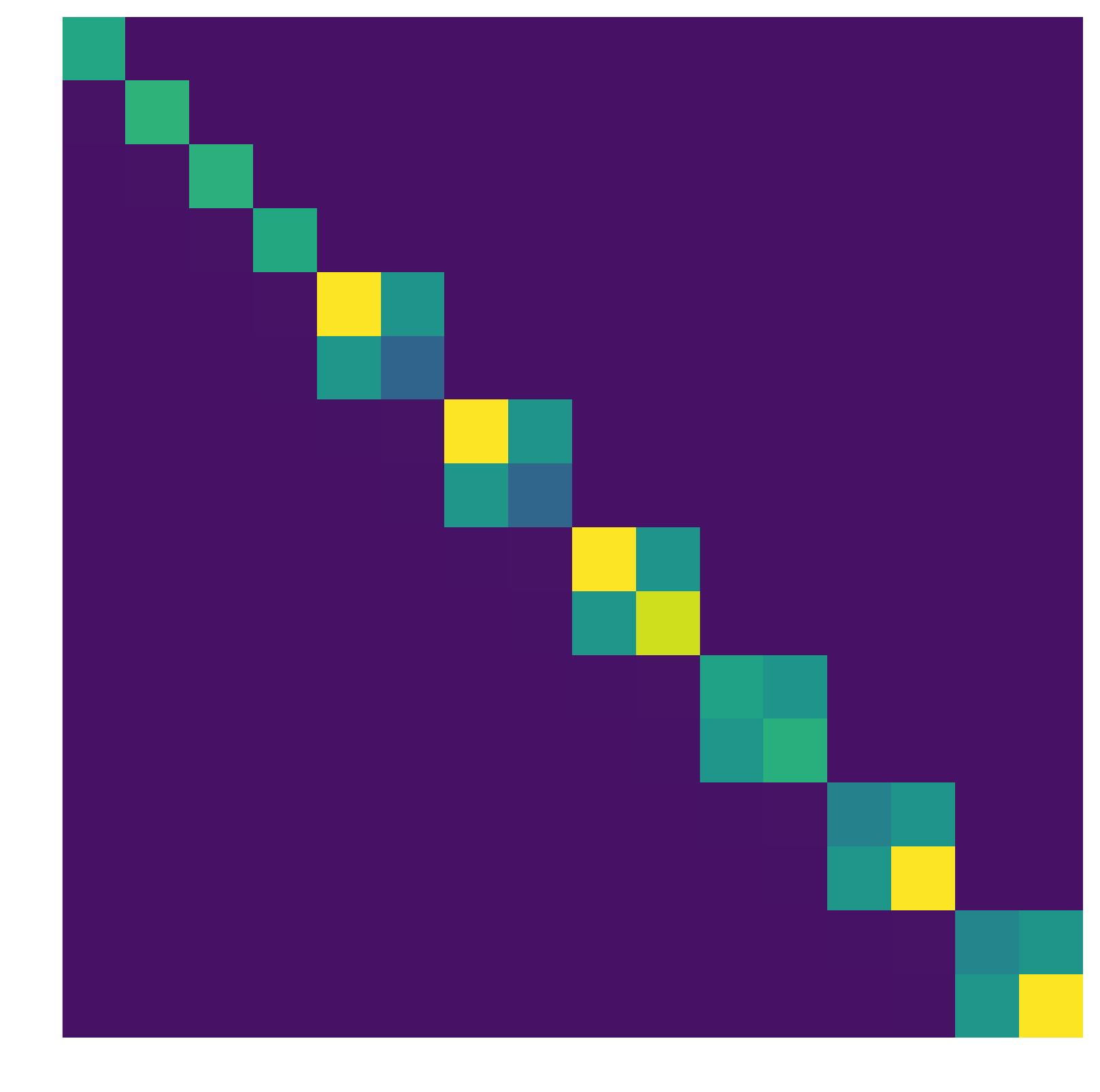}
  \caption{The Hessian (left), its $\tilde{H}^2$ approximation (middle)
  and its $\tilde{H}^1$ approximation (right) for a mixture of $N=4$ sources. %
  The fourth order tensors are reshaped into matrices of size $N^2 \times N^2$ for visualization purpose. The first $N$ rows correspond to the terms $\varepsilon_{ii}$, the following are arranged by pairs: $\varepsilon_{1,2}, \varepsilon_{2,1}, \varepsilon_{1,3}, \varepsilon_{3,1}\cdots\varepsilon_{N-1, N}, \varepsilon_{N, N-1}$. 
  This arrangement shows the block-diagonal structure of the approximations, the dark purple color corresponding to zero coefficients. 
  $\tilde{H}^2$ is $H$ stripped from any off-block coefficient, and $\tilde{H}^1$ slightly differs from $\tilde{H}^2$ on its diagonal.}
  \label{fig:shape}
\end{figure}

The Hessian approximations are discussed on the basis of the following moments:
\begin{equation}
\left\{
\begin{aligned}
  \hat{h}_{ijl} & =
    \hat{E}[\psi_i'(y_i)y_jy_l] \textrm{ , for }1 \leq i, j, l \leq N \\
  \hat{h}_{ij} & =
    \hat{E}[\psi_i'(y_i)y_j^2] \textrm{ , for }1 \leq i, j \leq N \\
    \hat{h}_{i} & =
    \hat{E}[\psi_i'(y_i)] \textrm{ , for }1 \leq i \leq N \\
    \hat{\sigma}_{i}^2 & =
    \hat{E}[y_i^2] \textrm{ , for }1 \leq i \leq N \\
\end{aligned}
\right.
\enspace .
\label{eq:hquantities}
\end{equation}
Hence, the true relative Hessian is $H_{i j k l} = \delta_{i l} \delta_{j k} + \delta_{ik} \hat{h}_{ijl}$.
A first approximation of $H$ consists in replacing $\hat{h}_{ijl}$ by $\delta_{jl} \hat{h}_{ij}$. We denote that approximation by $\tilde{H}^2$:
\begin{equation}
\tilde{H}^2_{ijkl} = \delta_{il}\delta_{jk} + \delta_{ik} \delta_{jl} \hat{h}_{ij} \enspace .
\label{eq:h2approx}
\end{equation}
A second approximation, denoted $\tilde{H}^1$, goes one step further and replaces $\hat{h}_{ij}$ by $\hat{h}_i \hat{\sigma}_j^2$ for $i \neq j$:
\begin{equation}
\left\{
\begin{aligned}
  \tilde{H}^1_{ijkl} & =
    \delta_{i l} \delta_{j k} + \delta_{ik} \delta_{jl}\hat{h}_i \hat{\sigma}_j^2\textrm{ , for }i \neq j\\
  \tilde{H}^1_{iiii} & = 1 + \hat{h}_{ii}
\end{aligned}
\right.
\enspace .
\label{eq:h1approx}
\end{equation}
Those two approximations are illustrated on Fig~\ref{fig:shape}.
A key feature is that both approximations are \emph{block diagonal}. Denoting $\tilde H$ for either $\tilde H^1$ or $\tilde H^2$, we note that, for $i \neq j$, the only non-zero coefficients in $\tilde{H}_{ijkl}$ are for $(k,l) = (i,j)$ and $(k,l)=(j,i)$.  The coefficients $\tilde{H}_{ijji}$ are equal to $1$.

When the signals are independent, $\hat{h}_{ijl} = \delta_{jl} \hat{h}_{ij} = \delta_{jl} \hat{h}_i \hat{\sigma}_j^2$ asymptotically in $T$ for $i \neq j$. 
This, together with $\hat{h}_{iii} = \hat{h}_{ii}$, means that \emph{the two approximations asymptotically match the true Hessian if the signals are independent}. In particular, if an iterative algorithm converges to a solution on a problem where the ICA mixture model holds, the Hessian approximations get very close to the true Hessian of the objective function.

Away from convergence or if the ICA mixture model does not hold, one cannot expect those approximations to be very accurate.  This is why we use them only as a \emph{preconditioners} for our algorithm.  They enjoy two properties which are critical in that respect, being fast to compute and easy to invert.  
Indeed, computing $\tilde{H}^1$ or $\tilde{H}^2$ is less costly than computing $H$. Evaluating $\tilde{H}^2$ requires the computation of $\hat{h}_{ij}$ for all $(i, j)$, which is of order $\Theta(N^2 \times T)$.  Obtaining $\tilde{H}^1$ is even faster, because it only requires the computation of the $\hat{h}_i$ and $\hat{\sigma}_i$, that is, $\Theta(N^1 \times T)$.
Furthermore, the  $\tilde H$ approximations are not only sparse: their block diagonal structure also allows $\tilde{H}^{-1}G$ to be computed quickly in close form.  
Indeed, defining $a_{ij} = \tilde{H}_{ijij}$, elementary linear algebra shows that
\begin{equation}
[\tilde{H}^{-1}G]_{ij} 
=
\frac
{a_{ji}G_{ij} - G_{ji}}
{a_{ij}a_{ji}-1} 
\qquad
\textrm{ for }i \neq j .
\end{equation}
Hence, computing $\tilde{H}^{-1}G$ has complexity $\Theta(N^2)$.


\subsection{Regularization of Hessian Approximations}

Like the true Hessian, the Hessian approximations have no reason to be positive definite. This means that we have to set up a regularization procedure.

That can be done at little cost since the two Hessian approximations can be diagonalized in close form by diagonalizing each of the $2 \times 2$ blocks. 
The smallest eigenvalue for the block $(i, j)$ is readily found to be: 
\begin{equation}
\lambda_{ij} = \frac{1}{2}( a_{ij} + a_{ji} - \sqrt{(a_{ij} -a_{ji})^2 + 4}) \enspace ,
\label{eq:eig}
\end{equation}
with, again, $a_{ij} = \tilde{H}_{ijij}$, for either $\tilde{H} = \tilde{H}^1$ or $\tilde{H} = \tilde{H}^2$.

Based on this, we propose the simple regularization procedure detailed in Algorithm~\ref{algo:reg}: the blocks with positive and large enough eigenvalues are left untouched, while the other blocks have their spectrum shifted so that their smallest eigenvalue is equal to a prescribed minimum value $\lambda_{min}$.

\begin{algorithm}[tb]
\SetKwInOut{Input}{Input}
\SetKwInOut{Output}{Output}
 \Input{Eigenvalue threshold $\lambda_{min} >0$, approximate Hessian $\tilde{H}$ ($\tilde{H}^1$ or $\tilde{H}^2$)}
 \For{Each pair $(i, j)$}{
  Compute $\lambda_{ij}$ using~\eqref{eq:eig}\;
  \If{$\lambda_{ij} < \lambda_{min}$}{
  Add  $(\lambda_{min} - \lambda_{ij})I_2$ to the block $(i, j)$ of $\tilde{H}$ \;
  }
  }
 \Output{Regularized $\tilde{H}$}
 \caption{Regularization procedure}
 \label{algo:reg}
\end{algorithm}

\section{Preconditioned ICA for Real Data}
\label{sec:prec}

Quasi-Newton methods attempt to estimate the local curvature of the objective function without explicitly computing its Hessian~\cite{nocedal1999optim}. Indeed, popular methods such as DFP~\cite{davidon1991variable, fletcher1970new, fletcher1963rapidly} or BFGS~\cite{broyden1970convergence,fletcher1970new,goldfarb1970family,shanno1970conditioning} build an approximation of the Hessian using solely function and gradient evaluations performed during optimization.

The popular L-BFGS~\cite{byrd1995limited} algorithm is used in many practical applications and obtains good results on a wide variety of problems.
Rather than storing all the updates of the Hessian approximation leading to a dense matrix potentially too big to fit in memory like BFGS does, L-BFGS only stores the last $m$ updates and then relies on a recursive inversion algorithm. The integer $m$ is referred to as the memory parameter. The algorithm starts from an initial guess for the Hessian which is easily invertible, and builds on it by adding low rank updates from the $m$ previous iterates and gradient evaluations. For lack of a better choice, vanilla L-BFGS uses a multiple of the identity matrix for the initial Hessian guess.  However, if some Hessian approximation is available, it could be used instead.  This is tantamount to preconditioning the problem with the said Hessian approximation~\cite{jiang2004preconditioned}.

The Hessian approximations $\tilde H$ provide us with a very effective preconditioning, as shown below in Sec.~\ref{sec:expe}, resulting in the `Preconditioned ICA for Real Data' (Picard) algorithm. 
Picard exploits the Hessian approximations to initialize the recursive formula of L-BFGS. It is summarized in algorithms~\ref{algo:lbfgs} and~\ref{algo:twoloop}.
We use the same notations as in \cite{nocedal1999optim}:
$y_i=G_i - G_{i-1}$, $s_i=a_i p_i$ (this is the ``relative'' update of the unmixing matrix between two iterations) and $\rho_i=1 / \langle s_i \lvert y_i \rangle$. 
As in standard L-BFGS algorithm, the search direction $p_k$ is computed using recursive algorithm with two \emph{for} loops, however the initial guess for the Hessian is here set to $\tilde{H}_k$. 
\begin{algorithm}[tb]
\SetKwInOut{Input}{Input}
\SetKwInOut{Output}{Output}
\Input{Mixed signals $X$, initial unmixing matrix $W_0$, memory size $m$, number of iterations $K$}
 \For{k=0,1,\dots,K}{
  Set $Y = W_k X$\;
  Compute the relative gradient $G_k$ using \eqref{eq:relatgrad}\;
  Compute Hessian approximation $\tilde{H}_k$ using \eqref{eq:h2approx} or \eqref{eq:h1approx}\;
  Regularize $\tilde{H}_k$ using algorithm \ref{algo:reg}\;
  Compute the search direction $p_k = - (\tilde{H}_k^{m})^{-1} G_k$ using L-BFGS formula in algorithm \ref{algo:twoloop}\;
  Compute the step length $\alpha_k$  using a line search\;
  Set $W_{k+1} = (I + \alpha_k p_k)W_k$ \;
  }
 \Output{$Y$, $W_k$}
 \caption{Preconditioned L-BFGS}
 \label{algo:lbfgs}
\end{algorithm}
\begin{algorithm}[tb]
\SetKwInOut{Input}{Input}
\SetKwInOut{Output}{Output}
 \Input{Current gradient $G_k$, Hessian approximation $\tilde{H}_k$, previous $s_i$, $y_i$, $\rho_i$ $\forall i \in \{k-m,\dots,k-1\}$.}
Set $q = -G_k$\;
 \For{i=k-1,\dots,k-m}{
	Compute $a_i = \rho_i \langle s_i \lvert q\rangle$ \;
    Set $q = q - a_i y_i$ \;
  }
  Set $r = \tilde{H}^{-1}_k q$ \;
  \For{i=k-m,\dots,k-1}{
  	Compute $\beta = \rho_i \langle y_i \lvert r \rangle$ \;
    Set $r = r +s_i(a_i - \beta)$ \;
  }
 \Output{$r=p_k$}
 \caption{Two loops recursion for L-BFGS using a preconditioner}
 \label{algo:twoloop}
\end{algorithm}

\subsection*{Line search}
\label{subsec:linesearch}

The algorithm relies on a line search procedure which aims at finding a good step size $\alpha$ at each iteration. In theory, the line search procedure has to enforce Wolfe conditions \cite{wolfe1969convergence,nocedal1999optim} in order to guarantee convergence.  The line search procedure proposed by Mor\'e and Thuente \cite{more1994line} is generally considered to be an efficient way to enforce such conditions. It is based upon cubic interpolation of the objective function in the direction of interest. Yet, for each candidate step size, one must compute the values of the objective function and of the gradient, which can be costly.

A simpler line search strategy is backtracking. If, for $\alpha=1$, the objective function is decreased, then that value is retained, otherwise the step size is divided by a factor of~$2$ and the process is repeated. This method only requires one evaluation of the likelihood at each step size, but it does not enforce Wolfe conditions.

In practice, backtracking is stopped when $\alpha$ becomes too small, which is an indication that the objective function has a pathological behavior in the search direction, since we rather expect values of the order of the ``Newton value''  $\alpha= 1$.  In the case of too many backtracking steps, resulting in too small a step size, the algorithm would not move much, and might get stuck for a long time in that problematic zone. Therefore, after a fixed number of failed backtracking step, the L-BFGS descent direction is deemed inefficient and we fall back to descending along the relative gradient direction, and reset the memory (we found that to happen quite infrequently in our experiments).

\section{Related work}
\label{sec:otherwork}

We compare our approach to the algorithms mentioned in section~\ref{sec:intro}. 
Some classical ICA algorithms such as FastICA~\cite{hyvarinen1999fast}, JADE~\cite{cardoso1993blind} or Kernel ICA~\cite{bach2002kernel} are not included in the comparison because they do not optimize the same criterion.

\subsection{Gradient descent}

The gradient is readily available and directly gives an update rule for a gradient descent algorithm:
\begin{equation}
    W \leftarrow \left( I - \alpha G \right) W \enspace ,
\label{eq:relupd}
\end{equation}
where $\alpha > 0$ is a step size found by line search or an annealing policy.
In the experiments, we used an oracle line-search: at each step, we find a very good step size using a costly line-search, but do not take into account the time taken, as if the sequence of best step sizes were readily available. This algorithm is referred to as "Oracle gradient descent".

\subsection{Infomax}
\label{subseq:infomax}

We now give a brief explanation on how the Infomax~\cite{bell1995information} algorithm actually runs.  It is a stochastic version of rule~(\ref{eq:relupd}): at each iteration of the algorithm, a relative gradient $G'$ is computed from a `mini-batch' of $T' \ll T$ randomly selected samples and a relative update $W \leftarrow (I - \alpha G') W$ is performed.

The stochasticity of Infomax has benefits and drawbacks. For a thorough review about what stochasticity brings, see~\cite{bottou2016optimization}. In summary, on the good side, stochasticity accelerates the first few passes on the full data because the objective starts decreasing after only one mini batch has been used, while for a full batch algorithm like the one presented above, it takes a full pass on the whole data to start making progress. Furthermore, if the number of samples is very large, computing the gradient using the whole dataset might be too costly, and then resorting to stochastic techniques is one way of coping with the issue.

Stochasticity, however, also comes with some disadvantages. The first one is that a plain stochastic gradient method with fixed batch size needs a very careful annealing policy for the learning rate to converge to a local minimum of the objective function. In practice, across iterations, the true gradient computed with the full set will not go to $0$, but instead will reach a plateau.

This is directly linked to the choice of the step size. If it is too small the algorithm will not make much progress, and if it is too large, the algorithm will become unstable. In fact, the level of the plateau reached by the gradient is proportional to the step size~\cite{bottou2016optimization}. Line search techniques are also unpractical, because one has only access to noisy realizations of the gradient and of the objective if one works only on a mini-batch of samples. In practice, the standard Infomax implementation relies on heuristics. It starts from a given step size $\alpha_0$, and decreases it by a factor $\rho$ if the angle between two successive search directions is greater than some constant $\theta$. That makes $3$ parameters that have to be set correctly, which may be problematic~\cite{montoya2017caveats}.

\subsection{Truncated Newton's method}
\label{subseq:trunc}

As explained above, direct Newton's method is quite costly. The so-called truncated Newton method~\cite{nocedal1999optim} manages to obtain directions similar to Newton's method at a fraction of the cost. The idea is to compute $H^{-1}G$ using the conjugate gradient method~\cite{nocedal1999optim} which does not require the construction of $H$ but only a mean of computing a Hessian-vector product $HM$. In the ICA problem, there is an efficient way to do so, a \emph{Hessian free product}. 
Indeed, using expression~\eqref{eq:hessian} of the true Hessian, one finds:
\[
  (HM)_{ij} = \sum_{k, l} H_{ijkl}M_{kl} = M_{ji} + \hat{E}[\psi_i'(y_i)y_j\sum_{l}M_{il}y_l]
\]
or in a matrix form:
\[
  HM = M^\top + \frac{1}{T}[\psi'(Y) \cdot (MY)] Y^\top
\]
where $\cdot$ is the element-wise matrix product. This computation comes at roughly the same cost as a gradient evaluation, its complexity being $\Theta(N^2 \times T)$.

The idea of truncated Newton's method is that instead of perfectly computing $H^{-1}G$ using conjugate gradient, one can stop the iterations before termination, at a given error level, and still obtain a useful approximation of Newton's direction.

This method is applied to ICA in~\cite{tillet2017infomax} where the authors also use a stochastic framework with variable batch size, speeding up the algorithm during the first steps. We did not implement such a strategy in order to have a fairer comparison.

One way of incorporating Hessian approximations in this method (not implemented in~\cite{tillet2017infomax}) is to use them once again as preconditioners for the linear conjugate gradient.  We found that this idea roughly halves the number of conjugate gradient iterations for a given error in solving $H^{-1}G$.

A difficulty arising with this method is the Hessian regularization. Because it avoids the computation of the Hessian, finding its smallest eigenvalue is not straightforward, and heuristics have to be used, like in~\cite{tillet2017infomax}. However, we do not want these hand tuned parameters to bias the algorithm comparison. Hence, in our implementation of the algorithm, we compute $H$ and its smallest eigenvalue $\lambda_{m}$ but  we do not include the associated cost in the timing of the algorithm. Then, we regularize $H$ by adding $ -2 \lambda_{m} Id$ to it if $\lambda_{m} < 0$.

These steps are summarized in algorithm~\ref{algo:truncnewt} in which the step  marked with a $(*)$ is not counted in the final timing.

\begin{algorithm}[htbp]
\SetKwInOut{Input}{Input}
\SetKwInOut{Output}{Output}
 \Input{Mixed signals $X$, initial unmixing matrix $W_0$, number of iterations $K$.}
Set $Y = W_0X$\;
 \For{k=0,1,\dots,K}{
  Compute relative gradient $G_k$ using \eqref{eq:relatgrad}\;
  $(*)$ Compute a regularization level $\lambda$ for $H_k$\;
  Compute the search direction $p_k = - (H_k+\lambda I)^{-1}G_k$ by preconditioned  conjugate gradient with regularized $\tilde{H_k}$\;
  Set $W_{k+1} = (I + \alpha_k p_k)W_k$ ($\alpha_k$ set by line search)\;
  Set $Y \leftarrow (I + \alpha_k p_k) Y$\;
  }
 \Output{$Y$, $W_k$}
 \caption{Truncated Newton's method}
 \label{algo:truncnewt}
\end{algorithm}

\subsection{Simple Quasi-Newton method}

The simplest way to take advantage of the Hessian approximations is to use them as replacement of $H$ in Newton algorithm. The descent direction is then given by $-\tilde{H}^{-1}G$.
We will refer to this as the simple quasi-Newton method, which is detailed in Algorithm~\ref{algo:elem}. Note that any Hessian approximation can be used as long as it is guaranteed to be positive definite. This optimization algorithm is used in~\cite{zibulevsky2003blind} with $\tilde{H}^2$ (however, the regularization technique differs from our implementation), and in~\cite{palmer2012amica} with $\tilde{H}^1$.

In the experiments, we refer to this algorithm as ``Simple quasi-Newton H2'' and ``Simple quasi-Newton H1'' where we respectively use $\tilde{H}^2$ and $\tilde{H}^1$ as approximations.

\begin{algorithm}[htbp]
\SetKwInOut{Input}{Input}
\SetKwInOut{Output}{Output}
 \Input{Mixed signals $X$, initial unmixing matrix $W_0$, number of iterations $K$.}
Set $Y = W_0X$\;
 \For{k=0,1,\dots,K}{
  Compute relative gradient $G_k$ using \eqref{eq:relatgrad}\;
  Compute Hessian approximation $\tilde{H_k}$ using \eqref{eq:h2approx} or \eqref{eq:h1approx}\;
  Regularize $\tilde{H}_k$ using algorithm \ref{algo:reg}\;
  Compute the search direction $p_k = - (\tilde{H_k})^{-1}G_k$\;
  Set $W_{k+1} = (I + \alpha_k p_k)W_k$ ($\alpha_k$ set by line search)\;
  Set $Y \leftarrow (I + \alpha_k p_k) Y$\;
  }
 \Output{$Y$, $W_k$}
 \caption{Simple quasi-Newton}
 \label{algo:elem}
\end{algorithm}

\subsection{Trust-region method}
Another way to proceed is to use a trust-region algorithm~\cite{nocedal1999optim}, rather than a line-search strategy. It is the idea proposed in~\cite{choi2007relative}, where $\tilde{H}^2$ is used to build a local quadratic approximation of the objective, and then minimization is done with a trust region update. In the experiments, we denote this algorithm by ``Trust region ICA''. For the experiments, we used a direct translation of the author's code in Python, which produces the same iterations as the original code.

\section{Experiments}
\label{sec:expe}

We went to great lengths to ensure a fair comparison of the above algorithms. We reimplemented each algorithm using the Python programming language.  Since the most costly operations are by far those scaling with the number of samples (\textit{i.e.} evaluations of the likelihood, score and its derivative, gradient, Hessian approximations and Hessian free products), we made sure that all implementations call the same functions, thereby ensuring that differences in convergence speed are caused by algorithmic differences rather than by details  of implementation.

Our implementations of Picard, the simple quasi-Newton method, the truncated Newton method and the trust region method are available online\footnote{https://github.com/pierreablin/faster-ica}.

\subsection{Experimental setup}

All the following experiments have been performed on the same computer using only one core of an Intel Core i7-6600U @ 2.6 GHz. For optimized numerical code we relied on Numpy~\cite{vanderwalt2011} using Intel MKL as the linear algebra backend library, and the numexpr package\footnote{ https://github.com/pydata/numexpr} to optimize CPU cache. It was particularly efficient in computing $\log\cosh(y_i(t) / 2)$ and $\tanh(y_i(t) / 2) \enspace \forall i, t$.

For each ICA experiment, we keep track of the gradient infinite norm (defined as $\max_{ij}\lvert G_{ij} \rvert $) across time and across iterations.  The algorithms are stopped if a certain predefined number of iterations is exceeded or if the gradient norm reaches a small threshold (typically $10^{-8}$).


Each experiment is repeated a certain number of times to increase the robustness of the conclusions. We end up with several gradient norm curves. On the convergence plots, we only display the median of these curves to maintain readability: half experiments finished faster and the other half finished slower than the plotted curve.

Besides the algorithms mentioned above, we have run a vanilla version of the L-BFGS algorithm, and Picard algorithm using $\tilde{H}^1$ and $\tilde{H}^2$.

\subsection{Preprocessing}
\label{subsec:preproc}
A standard preprocessing for ICA is applied in all our experiments, as follows.
Any given input matrix $X$ is first mean-corrected by subtracting to each row its mean value.
Next, the data are whitened by multiplication by a matrix inverse square root of the empirical covariance matrix   $C=\frac{1}{T}X X^\top$.
After whitening, the covariance matrix of the transformed signals is the identity matrix. In other words, the signals are decorrelated and scaled to unit variance. 

\subsection{Simulation study}

In this first study, we present results obtained on synthetic data. The general setup is the following: we choose the number of sources $N$, the number of samples $T$ and a probability density for each source. For each of the $N$ densities, we draw $T$ independent samples. Then, a random mixing matrix whose entries are normally distributed with zero mean and unit variance is created. The synthetic signals are obtained by multiplying the source signals by the mixing matrix and preprocessed as described in section~(\ref{subsec:preproc}).

We repeat the experiments $100$ times, changing for each run the seed generating the random signals.

We considerd $3$ different setups:
\begin{itemize}
\item \emph{Experiment A}: $T = 10000$ independent samples of $N=50$ independent sources. All sources are drawn with the same  density $p(x) = \frac{1}{2}\exp(-\lvert x \rvert)$.

\item \emph{Experiment B}: $T = 10000$ independent samples of $N=15$ independent sources. The 5 first sources have density  $p(x) = \exp(-\lvert x \rvert)$, the 5 next sources are Gaussian, and the 5 last sources have density $p(x) \propto \exp(-\lvert x^3 \rvert)$.

\item \emph{Experiment C}: $T = 5000$ independent samples of $N=40$ independent sources.  The $i$th  source has density $p_i = \alpha_i \mathcal{N}(0, 1) + (1-\alpha_i) \mathcal{N}(0, \sigma^2)$ where $\sigma = 0.1$ and  $\alpha_i$ is a sequence of linearly spaced values between $\alpha_1 =0.5$ and $\alpha_n =1$.

\end{itemize}

\begin{figure}
  \centering
  \includegraphics[clip,width=\columnwidth]{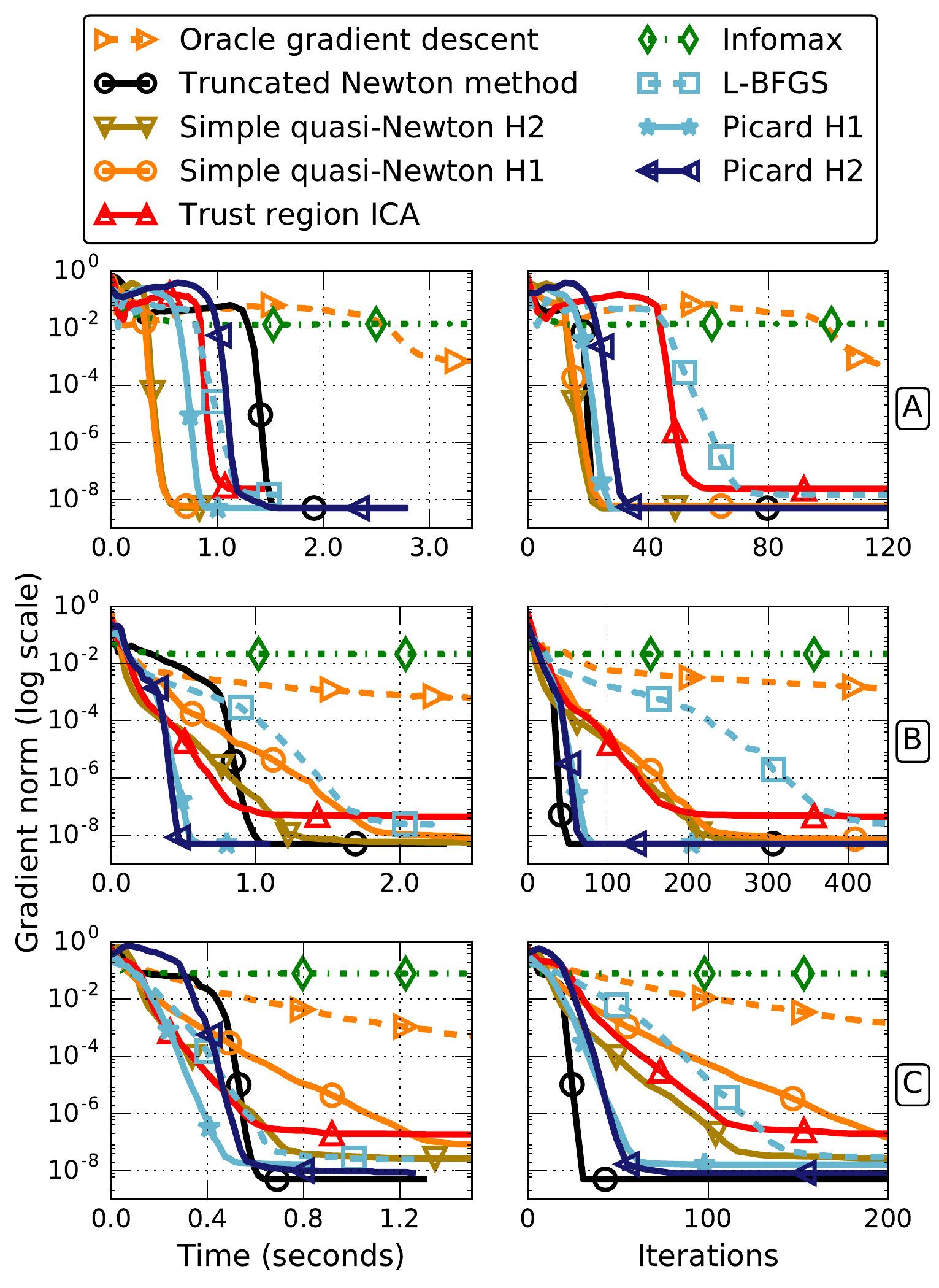}
    \caption{Comparison of the optimization algorithms on three synthetic experiments. Top: experiment A, middle: experiment B, bottom: experiment C. Left: infinity norm of the relative gradient as a function of time, right: same as a function of iterations (pass on the full set for Infomax). Solid lines correspond to algorithms informed of the approximate Hessians, dashed lines are their standard counterparts.}
     \label{fig:synthexp}
\end{figure}

In experiment A, the ICA assumption holds perfectly, and each source has a super Gaussian density, for which the choice $\psi = \tanh(\cdot / 2)$ is appropriate.
In experiment B, the first 5 sources can be recovered by the algorithms for the same reason. However, the next 5 sources cannot because they are Gaussian, and the last 5 sources cannot be recovered either because they are sub-Gaussian.
Finally, in experiment C, the mixture is identifiable but, because of the limited number of samples, the most Gaussian sources cannot be distinguished from an actual Gaussian signal. The results of the three experiments are shown in Figure~\ref{fig:synthexp}.

\subsection{Experiments on EEG data}

\begin{figure}
  \centering
  \includegraphics[clip,width=\columnwidth]{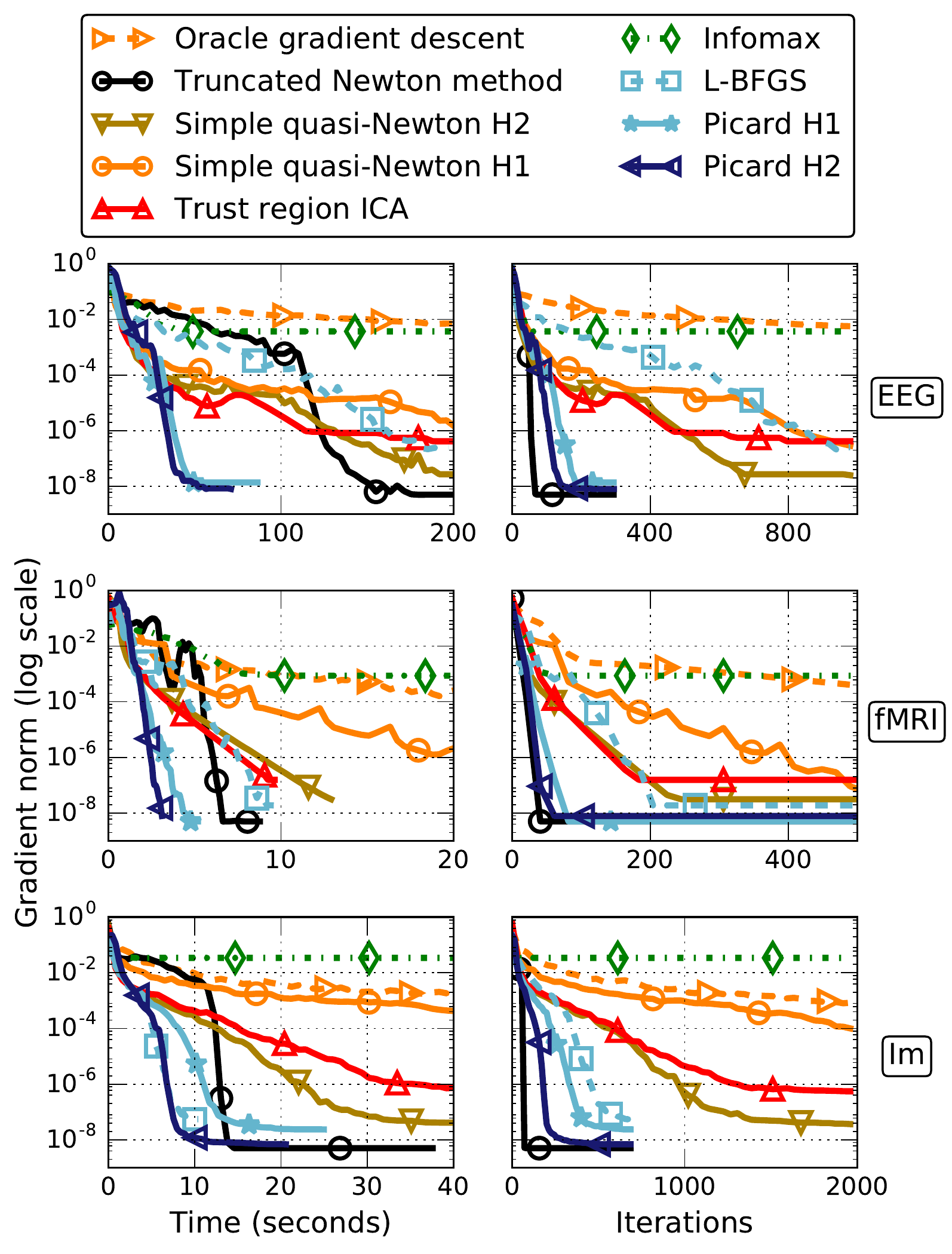}
    \caption{Comparison of the optimization algorithms on real data. Top: EEG dataset. Middle: fMRI dataset. Bottom: Image patches dataset. Left: infinity norm of the relative gradient w.r.t. time, right: same w.r.t. iterations (pass on the full data for Infomax). Solid lines correspond to algorithms informed of the approximate Hessian, dashed lines are their standard counterparts.}
     \label{fig:eegexpe}
\end{figure}

Our algorithms were also run on $13$ publicly available\footnote{https://sccn.ucsd.edu/wiki/BSSComparison} EEG datasets~\cite{delorme2012independent}. Each recording contains $n=71$ signals, and has been down-sampled by 4 for a final length of $T \simeq 75000$ samples.
EEG measures the changes of electric potential induced by brain activity. For such data, the ICA assumption does not perfectly hold. In addition, brain signals are contaminated by a lot of noise and artifacts. Still, it has been shown that ICA succeeds at extracting meaningful and biologically plausible sources from these mixtures~\cite{jung1998extended,Makeig30091997,delorme2012independent}.  Results are displayed on the top row of Figure~\ref{fig:eegexpe}.

\subsection{Experiments on fMRI data}

For those experiments, we used preprocessed fMRI data from the ADHD-200 consortium~\cite{adhd2012adhd}. We perform group ICA using the CanICA framework~\cite{varoquaux2010group} in Nilearn~\cite{abraham2014machine}. From the fMRI data of several patients, CanICA builds a signal matrix to which classical ICA is applied. We use that matrix to benchmark the algorithms. The problem is of size $N = 40$ and $T \simeq 60000$. See middle row of Figure~\ref{fig:eegexpe}.

\subsection{Experiments on natural images}

Given a grayscale image, we extract $T$ square patches (contiguous squares of pixels) of size $(s, s)$. Each patch is vectorized, yielding an $s^2\times T$ data matrix.  One may compute an ICA of this data set and see the columns of the mixing matrix $W^{-1}$ as features, or dictionary atoms, learned from random patches.
%
%

The ICA algorithms are run on a set of 100 natural images of open country\footnote{http://cvcl.mit.edu/database.htm}~\cite{oliva2001modeling},
using $T = 30000$ patches of side $8\times 8$ pixels, resulting in a $64 \times 30000$ data matrix. The patches are all centered and scaled so that their mean and variance equal 0 and 1 respectively before whitening as in section~\ref{subsec:preproc}.
Results are shown at the bottom of Figure~\ref{fig:eegexpe}. 




\subsection{Discussion}

On the first synthetic experiment, where the ICA mixture model holds, second order algorithms are all seen to perform well, converging in a handful of iterations. For this problem, the fastest algorithms are the simple quasi-Newton methods, which means that Picard does not improve significantly over the Hessian approximations $\tilde{H}^1$ or $\tilde{H}^2$.  This is expected since the Hessian approximations are very efficient when the ICA mixture model holds.

On the two other simulations, the ICA model is not identifiable because of the Gaussian signals. First order methods perform poorly. We can observe that for algorithms relying only on the Hessian approximations (simple quasi-Newton and trust-region ICA), the convergence speed is reduced. On the contrary, Picard and truncated Newton manage to keep a very quick convergence.
On those synthetic problems, it is not clear whether or not the greater accuracy of $\tilde{H}^2$ over $\tilde{H}^1$ justifies the added computation cost.

On EEG and fMRI data, Picard still converges quickly, in a fraction of the time taken by the other algorithms. For this problem, using $\tilde{H}^2$ for preconditioning leads to faster convergence than $\tilde{H}^1$.
The results are even more striking on images, where Picard, standard L-BFGS and truncated Newton converge in a few seconds while the other algorithms show a very slow linear convergence pattern.

On all experiments, truncated Newton's method converges in fewer iterations than Picard. This happens because it follows a direction very close to Newton's true direction, which is the direction each second order algorithm tries to mimic when the current iterate is close to the optimum. However, if we compare algorithms in terms of time, the picture is different: the reduced number of iterations does not make up for the added cost compared to Picard. 

\subsection{Complexity comparison of truncated Newton and preconditioned L-BFGS}

Truncated Newton's method uses the full information about the curvature. In our experiments, we observe that while this method converges in fewer iterations than Picard, it is slower in terms of CPU time. The speed of truncated Newton depends on many parameters (stopping policy for the conjugate gradient, regularization of $H$ and of $\tilde{H}$), so we propose a complexity comparison of this algorithm and Picard, to understand if the former might sometimes be faster than the latter.

Operations carried by the algorithms fall into two categories. First, there are operations that do not scale with the number of samples $T$, but only with the number of sources $N$. Regularizing the Hessian, computing $\tilde{H}^{-1}G$ and the L-BFGS inner loop are such operations.
The remaining operations scale linearly with $T$. Computing the score, its derivative, or evaluating the likelihood are all $\theta(N \times T)$ operations. The most costly operations are in $\Theta(N^2 \times T)$. They are: computing the gradient, computing $\tilde{H ^2}$, and for the truncated Newton method, computing a Hessian free product.

For the following study, let us reason in the context where $T$ is large in front of $N^2$, as it is the case for most real data applications. In that context, we do not count operations not scaling with $T$. This is a reasonable assumption on real datasets: on the EEG problem, these operations make for less than $1 \%$ of the total timing for Picard.

To keep the analysis simple, let us also assume that the operations in $\Theta(N \times T)$ are negligible in front of those in  $\Theta(N^2 \times T)$. When computing a gradient, the coefficients of $\tilde{H}^2$ or a Hessian free product, the costly operation in $\Theta(N^2 \times T)$ is numerically the same: it is the computation of a matrix product of the form $Y_1 Y_2^\top$ where $Y_1$ and $Y_2$ have the same shape as $Y$. With that in mind, we assume that each of these operations take the same time, $t_G$.

In order to produce a descent direction, Picard only needs the current gradient and Hessian approximations; the remaining operations do not scale with $T$. This means that each descent direction takes about $2 \times t_G$ to be found. This complexity is exactly the same as the simple quasi-Newton method.
On the other hand, truncated Newton requires the two same operations, as well as one Hessian-free product for each inner iteration of the conjugate gradient. If we denote by $N_{cg}$ the number of inner loops for the conjugate gradient, we find that truncated Newton's method takes $(2 + N_{cg}) \times t_G$ to find the descent direction.

Now, in our experiments, we can see that truncated Newton converges in about half as many iterations as Picard. Hence, for truncated Newton to be competitive, each of its iterations should take no longer than twice a Picard iteration.  That would require  $N_{cg}\leq 2$ but in practice, we observed that many more conjugate gradient iterations are needed (usually more than $10$) to provide a satisfying Newton direction. On the other hand, if the conjugate gradient algorithm is allowed to perform only $N_{cg}=2$ inner loops at each iteration, it results in a direction which is far from Newton's direction, drastically increasing the number of iterations required for convergence.

This analysis leads us to think that the truncated Newton's method as described in section~\ref{subseq:trunc} cannot be faster than Picard.

\subsection{Study of the control parameters of Picard}

Picard has four  control parameters: binary choice between two Hessian approximations, memory size $m$ for L-BFGS, number $n_{ls}$ of tries allowed for the backtracking line-search and regularization constant $\lambda_{min}$.
Our experiments indicate that $\tilde{H}^2$ is overall a better preconditioner for the algorithm, although the difference with $\tilde{H}^1$ can be small.

Through experiments, we found that the memory size had barely no effect in the range $3 \leq m \leq 15$. For a smaller value of $m$, the algorithm does not have enough information to build a good Hessian approximation. If $m$ is too large, the Hessian built by the algorithm is biased by the landscape explored too far in the past.

The number of tries for the line-search has a tangible effect on convergence speed. Similarly, the optimal regularization constant depends on the difficulty of the problem. However, on the variety of different signals processed in our experiments (synthetic, EEG, fMRI and image), we used the same parameters $m=7$, $n_{ls}= 10$ and $\lambda_{min}=10^{-2}$. 
As reported, those values yielded uniformly good performance.

\section{Conclusion}

While ICA is massively used across scientific domains, computation time for inference can be a bottleneck in many applications. The purpose of this work was to design a fast and accurate algorithm for maximum-likelihood ICA.

For this optimization problem, there are computationally cheap approximations of the Hessian. This leads to simple quasi-Newton algorithms that have a cost per iteration only twice as high as a gradient descent, while offering far better descent directions. Yet, such approximations can be far from the true Hessian on real datasets. As a consequence, practical convergence is not as fast as one can expect from a second order method.
Another approach is to use a truncated Newton algorithm, which yields directions closer to Newton's algorithm, but at a much higher cost per iteration.

In this work, we introduced the Preconditioned ICA for Real Data (Picard) algorithm, which combines both ideas. We use the Hessian approximations as preconditioners for the L-BFGS method. The algorithm refines the Hessian approximations to better take into account the true curvature. The cost per iteration of Picard is similar to the simple quasi-Newton methods, while providing far better descent directions. This was demonstrated, through careful implementation of various literature methods and extensive experiments over synthetic, EEG, fMRI and image data, where we showed clear gains in running time compared to the state-of-the-art.

\subsection*{Future work}

The algorithm presented in this article is developed with fixed score functions. It would be of interest to extend it to an adaptive score framework for the recovery of a broader class of sources. An option is to alternate steps of mixing matrix estimation and steps of density estimation, as is done in AMICA for instance. In preliminary experiments, such an approach was found to impair the convergence speed of the algorithm. More evolved methods have to be considered.

Second, the regularization technique presented here is based on a trial and error heuristic which has worked uniformly well on each studied dataset. Still, since the eigenvalues of the Hessian are driven by the statistics of the signals, a careful study might lead to more informed regularization strategies.

\section*{Acknowledgments}

This work was supported by the Center for Data Science, funded by the IDEX Paris-Saclay, ANR-11-IDEX-0003-02, and the European Research Council (ERC SLAB-YStG-676943).



\section*{Acknowledgment}

This work was supported by the Center for Data Science, funded by the IDEX Paris-Saclay, ANR-11-IDEX-0003-02, and the European Research Council (ERC SLAB-YStG-676943).

\bibliographystyle{IEEEtran}
\bibliography{bibliography}

\end{document}